\documentclass[journal,comsoc, onecolumn]{IEEEtran}
\ifCLASSINFOpdf
\else
\fi
\usepackage{amsmath}
\usepackage[cmintegrals]{newtxmath}
\usepackage{hyperref}

\usepackage{cite}
\usepackage{amsmath,amsfonts}
\usepackage{textcomp}
\usepackage{xcolor}
\usepackage{booktabs}

\usepackage{epsfig}
\usepackage{latexsym}
\usepackage{euscript}
\usepackage{multirow}
\usepackage{subcaption}
\usepackage{graphics,eepic,epic,psfrag}

\usepackage{textcomp}
\usepackage{algorithm}
\usepackage{algorithmic}
\usepackage{comment}

\usepackage{pifont}
\newcommand{\cmark}{\ding{51}}%
\newcommand{\xmark}{\ding{55}}%

\usepackage{graphicx, float, caption, array, multirow}

\usepackage[utf8]{inputenc}
\usepackage[english]{babel}

\renewcommand{\thetable}{\arabic{table}}
\newcolumntype{P}[1]{>{\centering\arraybackslash}p{#1}}

\def\BibTeX{{\rm B\kern-.05em{\sc i\kern-.025em b}\kern-.08em
		T\kern-.1667em\lower.7ex\hbox{E}\kern-.125emX}}

\begin{document}
%
\title{Improving Knowledge Distillation with\\ Teacher's Explanation}
%
%
%

\author{Sayantan~Chowdhury,~\IEEEmembership{Member,~IEEE,}
        Ben~Liang,~\IEEEmembership{Fellow,~IEEE,}
		Ali~Tizghadam,
        and~Ilijc~Albanese
\thanks{S. Chowdhury and B. Liang are with the Department
of Electrical and Computer Engineering, University of Toronto, Toronto, Ontario M5S 1A1, Canada (e-mail: sayantan.chowdhury@mail.utoronto.ca, liang@ece.utoronto.ca). A. Tizghadam and I. Albanese are
with TELUS Communications, Toronto, Ontario M5A 1P4, Canada (e-mail:
ali.tizghadam@telus.com, ilijc.albanese@telus.com).}

}

\maketitle

\begin{abstract}
Knowledge distillation (KD) improves the performance of a low-complexity student model with the help of a more powerful teacher. The teacher in KD is a black-box model, imparting knowledge to the student only through its predictions. This limits the amount of transferred knowledge. In this work, we introduce a novel Knowledge Explaining Distillation (KED) framework, which allows the student to learn not only from the teacher's predictions but also from the teacher's explanations. We propose a class of superfeature-explaining teachers that provide explanation over groups of features, along with the corresponding student model. We also present a method for constructing the superfeatures. We then extend KED to reduce complexity in convolutional neural networks, to allow augmentation with hidden-representation distillation methods, and to work with a limited amount of training data using chimeric sets. Our experiments over a variety of datasets show that KED students can substantially outperform KD students of similar complexity.
\end{abstract}


%
\IEEEpeerreviewmaketitle

\section{Introduction}
\label{sec:intro}

The computational complexity of machine learning often hinders its deployment in devices with limited hardware capability. Knowledge distillation (KD) addresses this problem by designing low-complexity student models that are trained utilizing the information from more powerful teacher models. In the conventional setting of KD \cite{Hinton15}, the student is trained with true labels as well as the teacher's predictions smoothed by a temperature parameter. This helps the student generalize better and thus achieve superior performance on the test set. At high temperatures, KD has been shown to be equivalent to model compression proposed by \cite{Caruana06}. The success of Hinton's distillation has led to the development of various other distillation frameworks in the subsequent years, including logit-based approaches similar to Hinton's original KD \cite{Vapnik16, Vapnik15, Mirzadeh20, Yuan20, Fang21, Zhao22, Zhang18, Lan18, Huang22} and more complex techniques that also utilize the teacher's intermediate-layer outputs \cite{Romero14, Zagoruyko17, Tung19, Kim18, Tian20, Yang21, Park19, Ahn19, Heo19, Liu19, Shu21, Chen21, Mei21}.

However, learning from the teacher's final or intermediate-layer predictions only transfers a fraction of the teacher's knowledge to the student. A good teacher should not just dictate the prediction outcome, but also explain to the student how to predict. This motivates us to propose a new \textit{Knowledge Explaining Distillation (KED)} framework that utilizes teachers who transfer knowledge with both predictions and explanations. 

One may add explanation to the black-box teacher used in conventional KD, by applying existing methods of Interpretable AI \cite{Ribeiro16, Lundberg17, Kononenko14}. However, the explanation thus generated is sample-specific, and hence ultimately is not useful for training a student. Instead of explaining a black-box teacher, we consider new interpretable teachers that output the contribution of features toward the final prediction. Furthermore, an interpretable teacher that provides per-feature-based explanation will impose strict requirements on the dependency among the features and thus will achieve poor accuracy. Therefore, we propose to first group the features into superfeatures \cite{Ribeiro16, Chowdhury19, Jullum21} and generate explanation only on the superfeatures. We name the resultant teachers \textit{superfeature-explaining teachers}. We show that such teachers transfer more knowledge to the students than the black-box ones in regular KD and hence improve the student performance.

Our main contributions are summarized as follows:
\begin{itemize}
	\item We propose a novel KED framework using superfeature-explaining teachers. A teacher in KED provides soft predictions on the training samples as well as their associated explanations. The student has similar but scaled-down architecture as the teacher and learns from the teacher's predictions and explanations. Instead of using sample-specific explanation in the conventional interpretable AI, we construct teachers that are architecturally hardwired to provide consistent explanation over different samples. We also develop an algorithm that groups feature into the required superfeatures.
	\item We further enhance the core design of KED in multiple directions. First, we extend KED to efficiently perform distillation with unstructured datasets using convolutional neural networks (CNNs). The model complexity of CNNs can be greatly reduced by our proposed method compared with existing KD techniques. Furthermore, we show that KED is easily composable with the prevalent hidden-representation distillation methods and thus can further improve the performance of the student. In addition, when the available training data is limited, we show that the explanations of different samples can be combined to construct labels for out-of-distribution samples. Thus, we introduce the concept of KED student training on a chimeric set and show that it is an effective method for data augmentation.
	\item We conduct extensive experiments over a variety of datasets including MNIST, FashionMNIST, Unicauca, CIFAR10, CIFAR100, and Tiny Imagenet. In all cases, we observe significant improvement over KD. As an example, for CIFAR100, a black-box student with the VGG8 architecture achieves 68.72\% and 70.39\% accuracy without and with KD, respectively. With KED, a student of similar complexity achieves 73.50\% accuracy. We also study the effect of combining KED with hidden-representation distillation methods and the chimeric set, showing that the KED approach is flexible and effective.
	
\end{itemize}

The rest of this paper is structured as follows. A brief survey of existing literature on distillation and explanation is provided in Section \ref{sec:related}. The general mathematical framework of knowledge distillation is discussed in Section \ref{sec:preliminaries}. Section \ref{sec:distillation} presents the details of our proposed KED framework. Section \ref{sec:extn} discusses the extensions of KED for CNNs, composition with hidden-representation methods and the chimeric set. Our experimental results are summarized in Section \ref{sec:results}, followed by conclusion in Section \ref{sec:conclusion}.

\section{Related Works}
\label{sec:related}

\subsection{Knowledge Distillation} The idea of KD first appeared in \cite{Caruana06}, where the authors showed that it is possible to distill the knowledge of an ensemble of machine learning models to a single model by matching logits. Subsequently, \cite{Hinton15} popularized the technique to design low-complexity models. These models are often easy to deploy and provide faster inference. 

There have been several extensions to the original KD framework. The authors of \cite{Vapnik16} unified Hinton's KD with the privileged teacher framework of \cite{Vapnik15} from a generalization error perspective. The authors of \cite{Mirzadeh20} further proposed gradual KD through a teaching assistant. In \cite{Yuan20}, the authors argued that KD is equivalent to label smoothing regularization and proposed a teacher-free distillation framework where the student is self-taught or it learns from a designed regularization distribution. A data-free distillation method was introduced in \cite{Fang21} based on contrastive model inversion, where a generative model was used to obtain synthetic data for distillation. The authors of \cite{Zhao22} presented decoupled KD, enabling efficient distillation for target class and non-target classes. This work also demonstrated the efficacy of logit-based distillation. In \cite{Zhang18} and \cite{Lan18}, the authors investigated online KD, where the teacher and the student learn together mutually from each other.

None of these works consider teachers that explain the predictions that are taught to the students. In contrast, our work shows that substantial learning improvement can be achieved by a teacher that explains.

\subsection{Hidden-Representation Distillation Methods} The hidden-representation distillation framework has been developed in parallel to KD. In this framework, the student attempts to match the teacher's intermediate layer outputs, instead of the teacher's logits or final inference outcomes as in KD. In \cite{Romero14}, the student is a thin model and its latent representation is aligned with the teacher's intermediate layer hints using a projector. In \cite{Zagoruyko17} and \cite{Tung19}, the dependency in \cite{Romero14} upon the projector is removed by attention transfer and similarity preservation over training samples. The authors of \cite{Kim18} proposed paraphrasing a complex teacher by extracting so-called factors from the hidden layers. The authors of \cite{Tian20} and \cite{Yang21} proposed distillation via contrastive representation and softmax regression, respectively. In \cite{Park19}, the authors explored the distillation of relational knowledge between different samples as characterized by the teacher. An information-theoretic knowledge transfer scheme was introduced in \cite{Ahn19} that maximizes mutual information between the teacher and the student. The authors of \cite{Heo19} investigated the distillation of activation boundaries produced by the hidden neurons where a student matches the teacher's boundaries instead of the output of the layers. A structured knowledge distillation framework was proposed in \cite{Liu19} for semantic segmentation. In \cite{Shu21}, the channel-wise probability maps are used for distillation so that the student learns the most salient parts of each channel for dense prediction. Cross-layer matching of hidden representations was studied in \cite{Chen21} and \cite{Mei21}, which transfers knowledge via connection paths between different stages of the teacher and the student network.

However, learning directly from the teacher's hidden layers may not be always effective. Since the student model usually has far lower complexity than the teacher, trying to match the teacher's hidden layers may distract from the student's final classification task. In contrast, KED constructs a teacher that offers explicit explanations on groups of features, and it allows a hierarchy of different numbers of superfeatures. Furthermore, a distinguishing feature of KED is that the explanations can be considered as Shapley values \cite{Shapley53} of a cooperative game. More importantly, as shown later, KED can be augmented with existing hidden-representation distillation methods in scenarios when they are effective.
 
\subsection{Interpretable AI} To design an explaining teacher, we must circumvent the general lack of interpretability in modern machine learning techniques. Interpretable AI deals with this problem by providing explanations for the model's predictions. This is often achieved through feature attribution. The authors of \cite{Ribeiro16} proposed local interpretable model-agnostic explanation (LIME), which provides insights about the classifier's perception for a given sample. The authors of \cite{Lundberg17} unified existing feature attribution methods in the Shapley value-based framework and introduced low complexity algorithms for computing approximate Shapley values. The cooperative game defined in \cite{Kononenko14} was generalized in \cite{Chowdhury19, Jullum21} using the set of superfeatures as players and thus obtaining superfeature-based explanation. However, as detailed in Section \ref{subsec:lim}, the explanations generated by the above methods are local to specific samples and hence not conducive to knowledge transfer in distillation. A new approach is needed to integrate interpretable AI and knowledge distillation in our KED framework.

\section{Preliminaries on Knowledge Distillation}
\label{sec:preliminaries}

We first outline the general concept of knowledge distillation. We will set up the basic mathematical notations that will be used throughout this paper. Consider a classification problem given a labeled training dataset $\mathcal{D}$. Let $\mathcal{X}$ be the set of $d$-dimensional training samples and $\mathcal{Y}$ be the set of classes with $|\mathcal{X}|=K$ and $|\mathcal{Y}|=C$. Let $\mathbf{x}\in \mathcal{X}$ denote the feature vector of a sample and $y\in\mathcal{Y}$ denote its label. Then we have $\mathcal{D}=\{(\mathbf{x}, y)\}_{\mathbf{x}\in\mathcal{X}}$. In the classical, i.e., no distillation, setting, a classifier $g$ minimizes the following categorical crossentropy loss function:
\begin{equation}
\mathcal{L}(g)=\mathbb{E}_\mathcal{D}[\mathcal{L}_\text{ce}(y,g(\mathbf{x}))].
\label{eq:nd}
\end{equation}

The goal of KD is to design a low-complexity \textit{student model} with the help of a more powerful \textit{teacher model}. Let $f$ be the teacher that provides predictions for the student $g$. The student $g$ is trained with the hard labels from the dataset $\mathcal{D}$ and some soft labels supplied by the teacher $f$ \cite{Hinton15}. The soft labels are the teacher's logits passed through a softmax function with a temperature parameter, $T$, or equivalently the teacher's predictions $f(\mathbf{x})$ passed through a function $\sigma_T(\cdot)$ defined as follows. Let $\{p_i: i\in\{1,...,C\}\}$ denotes the teacher's predicted probability distribution over $C$ classes. Then, $\sigma_T(\cdot)$ is given by
\begin{equation}
\sigma_T(p_i)=\frac{\exp(\log p_i/T)}{\sum_{j=1}^{C}\exp(\log p_j/T)}, \;\forall i\in\{1,...,C\}.
\label{eq:softmax_T}
\end{equation}
The student minimizes the following loss function during training:
\begin{equation}
\begin{split}
\mathcal{L}(g)=(1-\lambda)&\mathbb{E}_\mathcal{D}[\mathcal{L}_\text{ce}(y, g(\mathbf{x}))]\\
&+ T^2\lambda\mathbb{E}_\mathcal{D}[\mathcal{L}_\text{kl}(\sigma_T(f(\mathbf{x})), \sigma_T(g(\mathbf{x})))],
\end{split}
\label{eq:kd}
\end{equation}  
where $y$ is the hard label, $\mathcal{L}_\text{ce}$ is the categorical crossentropy loss, and $\mathcal{L}_\text{kl}$ is the KL divergence loss. The second term is scaled by $T^2$ as prescribed in \cite{Hinton15}, and the loss weight $\lambda$ is a hyperparameter. The original KD approach has many successes in generating low-complexity models with superior performance \cite{Hinton15, Kim18, Mirzadeh20}. As seen in (\ref{eq:kd}), the teacher in KD remains a black-box model that provides only limited information to the student. In this work, our objective is to design a richer distillation framework that is more conducive to student learning, where the teacher provides explanations for its predictions in addition to the soft labels.

\section{Knowledge Explaining Distillation}
\label{sec:distillation}

In this section, we elaborate upon the proposed KED framework. We first discuss the limitation of existing explanation methods and the motivation for constructing superfeature-explaining teachers. Then, we describe the general architecture of such teachers and the corresponding student model, followed by the mathematical description of KED. Finally, we present an algorithm to group features into superfeatures.

\subsection{Limitation of Existing Explanation Methods}
\label{subsec:lim}

It is nontrivial to add explanation to the teacher in KD. For a sample $\mathbf{x}\in\mathcal{X}$ and a pretrained teacher $f$, the existing feature attribution-based explanation methods \cite{Kononenko14, Lundberg17} quantify the contribution of each feature to the final prediction $f(\mathbf{x})$. However, this explanation is local and specific to the sample $\mathbf{x}$, in the sense that the same feature value can be mapped to different explanations for different samples and thus the underlying relation between input features and explanation becomes one-to-many. Thus the target function between the features and the teacher's explanation obtained using these methods is simply undefined. Therefore, a student cannot estimate the target function from such an explanation in a distillation framework. This argument holds also for the superfeature-based explanation methods \cite{Ribeiro16, Chowdhury19, Jullum21}.

The limitation of existing explanation methods motivates us to develop more interpretable teachers instead of directly applying current methods to explain black-box KD teachers. To facilitate effective learning by students, interpretable teachers should be learning models hardwired to provide consistent explanations over different samples. In the next section, we will see that the function between the features and their explanation is built into the architecture of the proposed superfeature-explaining teachers. Therefore, this class of teachers is perfectly suitable for guiding the students in a distillation setting.

\subsection{Superfeature-Explaining Teachers}


\begin{figure}[t]
	\centering
	\includegraphics[width=8.5cm]{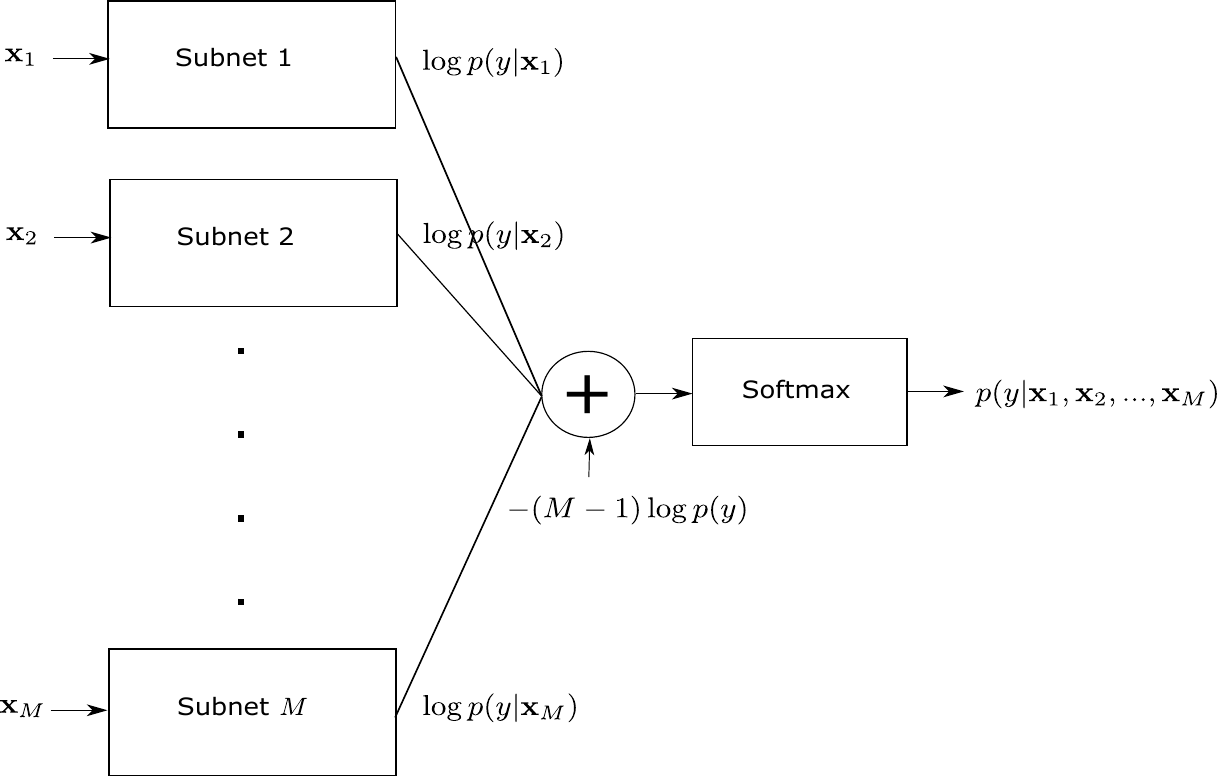}
	\caption{Architecture of superfeature-explaining teacher.}
	\label{fig:sems}
\end{figure}

\begin{figure}[t]
	\centering
	\includegraphics[width=8.5cm]{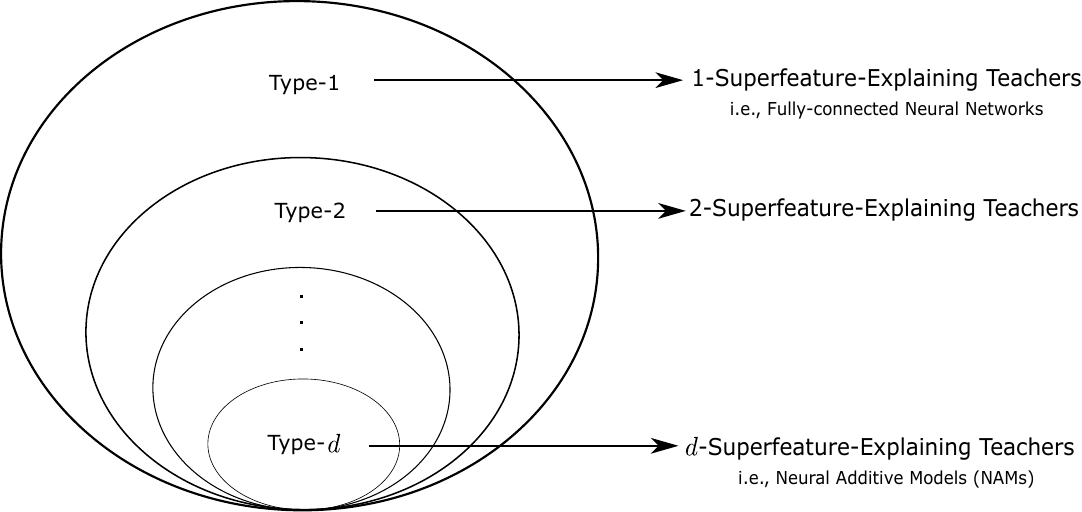}
	\caption{Hierarchy of superfeature-explaining teachers.}
	\label{fig:hierarchy}
\end{figure}

\begin{figure*}[t]
	\centering
	\includegraphics[width=18cm]{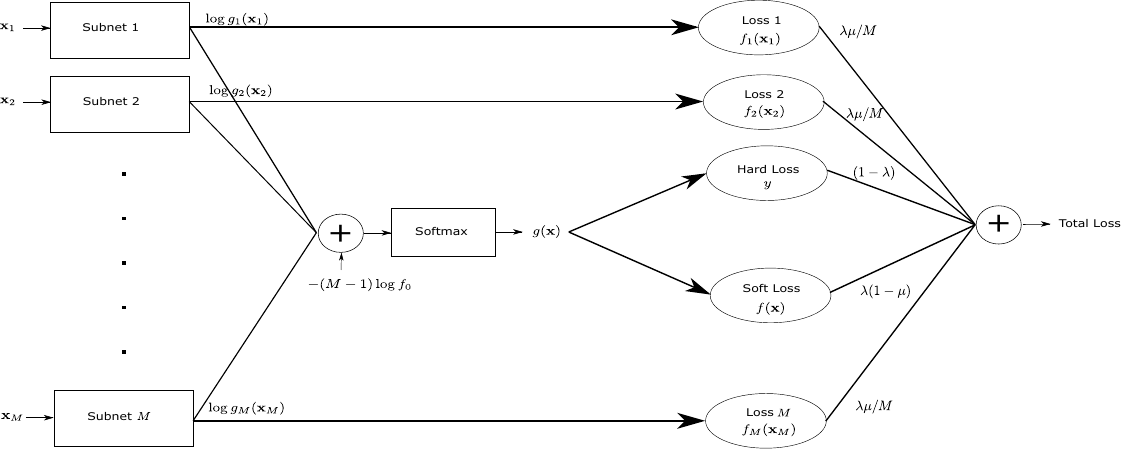}
	\caption{Student in KED.}
	\label{fig:shapdist}
\end{figure*}

In one extreme, the black-box teacher in the original KD does not provide any explanation to the student. In the other extreme, however, a teacher that provides an explanation for each feature may be too restrictive. Therefore, a more general approach is to obtain an explanation for groups of features. In this work, we consider superfeature-based explanation for constructing an interpretable teacher in a distillation framework. We define superfeatures as disjoint groups of features. Let $\mathbf{x}\in \mathcal{X}$ be the feature vector of a sample. Then, we can group the features in $\mathbf{x}$ as a set of superfeatures $\{\mathbf{x}_1,...,\mathbf{x}_M\}$ where $\mathbf{x}_m,\; 1\leq m\leq M,$ is the $m$-th superfeature. We will defer the discussion on how to group features to Section \ref{subsec:algo}.

Given an input $\mathbf{x}$, any classifier's job is to estimate the distribution $p(y|\mathbf{x})$ for each particular class label $y\in\mathcal{Y}$. To motivate our design, let's first suppose we have $M$ independent superfeatures. Then, by Bayes' rule, we have
\begin{equation}
p(y|\mathbf{x})=\frac{\frac{1}{p(y)^{M-1}}\prod_{m=1}^{M}p(y|\mathbf{x}_m)}{\sum_{y}\frac{1}{p(y)^{M-1}}\prod_{m=1}^{M}p(y|\mathbf{x}_m)},
\label{eq:sf}
\end{equation} 
where $p(y)$ is the prior of class $y$. Thus $p(y|\mathbf{x}_m)$ may be viewed as the contributions of the superfeature $\mathbf{x}_m$ to the prediction $p(y|\mathbf{x})$. We will use $p(y|\mathbf{x}_m), \forall m$, as the teacher's explanation for its prediction $p(y|\mathbf{x})$.

Figure \ref{fig:sems} illustrates the architecture of a superfeature-explaining teacher with $M$ superfeatures. This teacher estimates the contribution of each superfeature using a subnet, which is a neural network with a softmax layer at the output. We compute the logarithm of the softmax outputs from all the subnets and sum them up to produce the total logit adjusted by the prior. An ultimate softmax function is applied to this total logit in order to obtain the final prediction $p(y|\mathbf{x})$. Both the teacher's final prediction $p(y|\mathbf{x})$ and its superfeature explanation $p(y|\mathbf{x}_m), \forall m$, will be used as inputs to train the student model.

We remark that there is a hierarchy of superfeature-explaining teachers according to the number of superfeatures. A black-box teacher may be viewed as having only one superfeature, i.e., the set of all features. On the other hand, the generalized additive model, e.g., NAM \cite{Agarwal21} may be viewed as having $d$ superfeatures, i.e., each feature is a superfeature. The hierarchy of superfeature-explaining teachers is shown in Figure \ref{fig:hierarchy}. At the top of this hierarchy sits type-1 teachers or the black-box teachers that are the most unrestricted. Type-1 teachers do not assume independence between their input features and therefore theoretically can estimate any distribution that can be estimated by the types below it. At the bottom of this hierarchy is type-$d$ teachers that are the most restricted as they assume independence of all the features. As we go down the hierarchy, we see more interpretability of the teacher model at the expense of a drop in its prediction accuracy. The proposed KED in this work can be applied to any type. In particular, for type-1 teachers, our method reduces to Hinton's KD.

Interestedly, an alternative interpretation of superfeature-explaining teachers can be obtained using cooperative games. A cooperative game $\mathcal{G}=\langle\mathcal{N}, v\rangle$ is defined by a set of players, $\mathcal{N}$, and a value function, $v: 2^\mathcal{N}\rightarrow \mathbb{R}$. The Shapley value \cite{Shapley53} of each player measures the average marginal contribution of the player to the game's outcome. Let $\mathcal{N}$ be the set of superfeatures in KED, and we define the value function for any coalition $\mathcal{S}$ as 
\begin{equation}
v(\mathcal{S})=\sum_{m\in\mathcal{S}}\log p(y|\mathbf{x}_m)-(M-1)\log p(y).
\label{eq:value} 
\end{equation}
Note that the value of the grand coalition in this game is the total logit, i.e., 
\begin{equation}
v(\mathcal{N})=\sum_{m=1}^M\log p(y|\mathbf{x}_m)-(M-1)\log p(y).
\end{equation} 
For any sample $\mathbf{x}$, the Shapley value of the $m$-th superfeature can be obtained as 
\begin{equation}
q_m(\mathbf{x}_m, y) = \sum_{\mathcal{S}\,\subseteq\, \mathbf{x} \, \setminus \, \{ \mathbf{x}_m    \}} \frac{|\mathcal{S}|!(M-|\mathcal{S}|-1)!}{M!} [v(\mathcal{S}\cup \{\mathbf{x}_m\})-v(\mathcal{S})],
\label{eq:shapley} 
\end{equation}
where $v(\mathcal{S}\cup \{\mathbf{x}_m\})$ indicates the value of the game when the $m$-th superfeature is included in the coalition $\mathcal{S}$. Substituting (\ref{eq:value}) in (\ref{eq:shapley}), the Shapley value of the $m$-th superfeature in this game is given by 
\begin{equation}
\begin{split}
q_m(\mathbf{x}_m, y)&=\sum_{\mathcal{S}\,\subseteq\, \mathbf{x} \, \setminus \, \{ \mathbf{x}_m    \}} \frac{|\mathcal{S}|!(M-|\mathcal{S}|-1)!}{M!}\log p(y|\mathbf{x}_m)\\
&=\log p(y|\mathbf{x}_m).
\end{split}
\end{equation}
Hence, we establish that the outputs of the superfeature-explaining teacher are Shapley values. 

\subsection{Student in KED}
\label{subsec:ked}

In the KED framework, a type-$M$ student $g$ is trained with the hard labels from $\mathcal{D}$, and the soft labels and explanations provided by a type-$M$ teacher $f$. As in the original KD, the soft labels are the teacher's predictions $f(x)$ passed through function $\sigma_T(\cdot)$ as defined in (\ref{eq:softmax_T}). Let $f_m(\mathbf{x}_m), \; 1\leq m\leq M,$ be the teacher's explanation for the $m$-th superfeature. The student outputs $g_m(\mathbf{x}_m)$, which is matched against $f_m(\mathbf{x}_m)$ to compute the loss. We denote the teacher's as well as the student's prior by $f_0$. We further apply another temperature parameter $\tau$ to the teacher's explanation passing through function $\sigma_\tau(\cdot)$, which is as defined in (\ref{eq:softmax_T}) with $T$ replaced by $\tau$. 

The student's total loss is measured as a weighted combination of individual loss against the hard labels, the soft labels, and the explanations. Therefore, the student's loss function consists of three terms. The first term is the categorical crossentropy loss between the student's predictions $g(\mathbf{x})$ and the hard labels $y$. The second term is the KL divergence loss between the student's soft label predictions $\sigma_T(g(\mathbf{x}))$ and the teacher's soft labels $\sigma_T(f(\mathbf{x}))$ at temperature $T$. The third term is the average KL divergence loss over $M$ superfeatures between the student's outputs $\sigma_\tau(g_m(\mathbf{x}_m))$ and the teacher's explanation $\sigma_\tau(f_m(\mathbf{x}_m)), \; 1\leq m\leq M,$ at temperature $\tau$. Thus we write the student's total loss as
\begin{equation}
\begin{split}
\mathcal{L}(g)&=(1-\lambda)\mathbb{E}_\mathcal{D}[\mathcal{L}_\text{ce}(y, g(\mathbf{x}))]\\
& + T^2\lambda(1-\mu)\mathbb{E}_\mathcal{D}[\mathcal{L}_\text{kl}(\sigma_T(f(\mathbf{x})), \sigma_T(g(\mathbf{x})))]\\
& + \frac{\tau^2\lambda\mu}{M}\sum_{m=1}^{M}\mathbb{E}_\mathcal{D}[\mathcal{L}_\text{kl}(\sigma_\tau(f_m(\mathbf{x}_m)), \sigma_\tau(g_m(\mathbf{x}_m)))].
\end{split}
\label{eq:kd_loss}
\end{equation}  
The loss weights $\lambda$ and $\mu$ are two hyperparameters. Figure \ref{fig:shapdist} illustrates the student in the KED framework.
\subsection{Constructing Superfeatures}
\label{subsec:algo}

One key aspect of superfeature-explaining teachers is the superfeatures themselves. If there is sufficient domain knowledge, the data can be collected in a way such that the groups of features naturally form independent superfeatures. However, this is not always possible in general machine learning tasks. An alternative strategy is to group strongly dependent features together into a superfeature. Here we use a Hessian-based measure of feature dependency. Referring to Figure \ref{fig:sems}, we define the total logit $z$ for a type-$M$ superfeature-explaining teacher as follows:
\begin{equation}
z=\sum_{m=1}^{M}\log p(y|\mathbf{x}_m)-(M-1)\log p(y).
\end{equation}
We see that $z$ is additively separable over the superfeatures $\mathbf{x}_1,..., \mathbf{x}_M$. Therefore, if two features $x_i$ and $x_j$ belong to two different superfeatures, the cross-partial derivatives of $z$ with respect to those two features should be zero, i.e.,
\begin{equation}
\frac{\partial^2 z}{\partial x_i \partial x_j}=\frac{\partial^2 z}{\partial x_j \partial x_i}=0, \quad \forall x_i\in \mathbf{x}_m, x_j\in\mathbf{x}_n, m\neq n.
\end{equation}
Thus, we measure the dependency between each pair of features by the entries of $d\times d$ Hessian matrix of $z$. Note that only the magnitude of Hessian entries is significant for determining dependency.

We note that the total logit $z\approx \log p(y|\mathbf{x})$. We first estimate $p(y|\mathbf{x})$ using a type-$1$ black-box teacher. Then, we approximate the expectation of the Hessian matrix, $\mathbf{\bar{H}}\approx\sum_{y\in\mathcal{Y}} \mathbb{E}[\nabla\otimes\nabla \log p(y|\mathbf{x})]$ over a random subset of training samples. Let $\mathbf{\bar{H}}^{}_\text{abs}$ denote the matrix of absolute values of the entries in $\mathbf{\bar{H}}$. We define the $d\times d$ pairwise dependency matrix as $\mathbf{W}=\mathbf{\bar{H}}^{}_\text{abs}+\mathbf{\bar{H}}_\text{abs}^T$. Thus the $(i,j)$ entry of this matrix is
\begin{equation}
\mathbf{W}_{ij}\approx\left|\sum_{y\in\mathcal{Y}}\mathbb{E}\left[\frac{\partial^2 z}{\partial x_i \partial x_j}\right]\right|+\left|\sum_{y\in\mathcal{Y}}\mathbb{E}\left[\frac{\partial^2 z}{\partial x_j \partial x_i}\right]\right|.
\end{equation}

Now we are ready to build a weighted undirected graph $G=(V,E)$ where $V$ is the set of vertices, i.e., $d$ features, $E$ is the edges between each pair of those features, and each edge is assigned a weight given by the matrix $\mathbf{W}$. Then, the grouping of features into approximately independent superfeatures corresponds to finding approximately independent communities in graph $G$. We may use any community detection method, e.g., Louvain \cite{Blondel08}, to find such superfeatures. Even though many practical problems require a large number of features and hence a large graph $G$, Louvain is a simple and fast algorithm that can detect communities efficiently in $O(|V|\log|V|)$ run time.

\begin{algorithm}[t]
	\caption{Algorithm for constructing superfeatures}
	\begin{algorithmic}[1]
		\renewcommand{\algorithmicrequire}{\textbf{Input:}}
		\renewcommand{\algorithmicensure}{\textbf{Output:}}
		\REQUIRE Resolution
		\ENSURE Set of superfeatures 
		\STATE Compute average Hessian matrix over a random subset of training samples, $\mathbf{\bar{H}}\approx\sum_{y\in\mathcal{Y}} \mathbb{E}[\nabla\otimes\nabla \log p(y|\mathbf{x})]$.
		\STATE Construct a matrix $\mathbf{\bar{H}}^{}_\text{abs}$ taking absolute values of entries in $\mathbf{\bar{H}}$.  
		\STATE $\mathbf{W}\leftarrow \mathbf{\bar{H}}^{}_\text{abs}+\mathbf{\bar{H}}^T_\text{abs}$
		\STATE $\text{diag}(\mathbf{W})\leftarrow 0$
		\STATE Set up a graph with weight matrix $\mathbf{W}$.
		\STATE Perform community detection using the Louvain algorithm with the given resolution.
		\STATE Record the communities, i.e., the set of superfeatures.
		\STATE \textbf{return} Set of superfeatures
	\end{algorithmic} 
	\label{algo:sf}
\end{algorithm}

Algorithm \ref{algo:sf} summarizes the above procedure for constructing the superfeatures. The detected communities constitute our set of superfeatures. Typically, we find a higher or lower number of superfeatures by tuning the resolution parameter of the Louvain method.
\section{Extensions: CNN, Hidden-Representation and Chimeric Set}
\label{sec:extn}

In this section, we explore three interesting extensions of the KED framework. For CNNs, our proposed technique can significantly improve the distillation performance. Furthermore, we show how KED can be augmented with hidden-representation distillation methods. Also, in scenarios with limited training data, the performance of KED can be enhanced using a chimeric set.

\subsection{KED for CNNs}
\label{subsec:cnns}

Since the pixels of the input images are usually unstructured, a CNN has to extract its own features. As mentioned in Section \ref{subsec:algo}, if we can design the features such that they form independent groups, then we do not need any posterior process of constructing superfeatures. Thus, for building a type-$M$ CNN-based superfeature-explaining teacher, it is enough to split the architecture into $M$ subnets at some intermediate layer so that during training, the backpropagation algorithm can construct the superfeatures automatically. As the explanations from $M$ subnets are combined at the output end using (\ref{eq:sf}), reducing the training loss is equivalent to promoting independence among the superfeatures.

\begin{figure}[t]
	\centering
	\includegraphics[width=8.5cm]{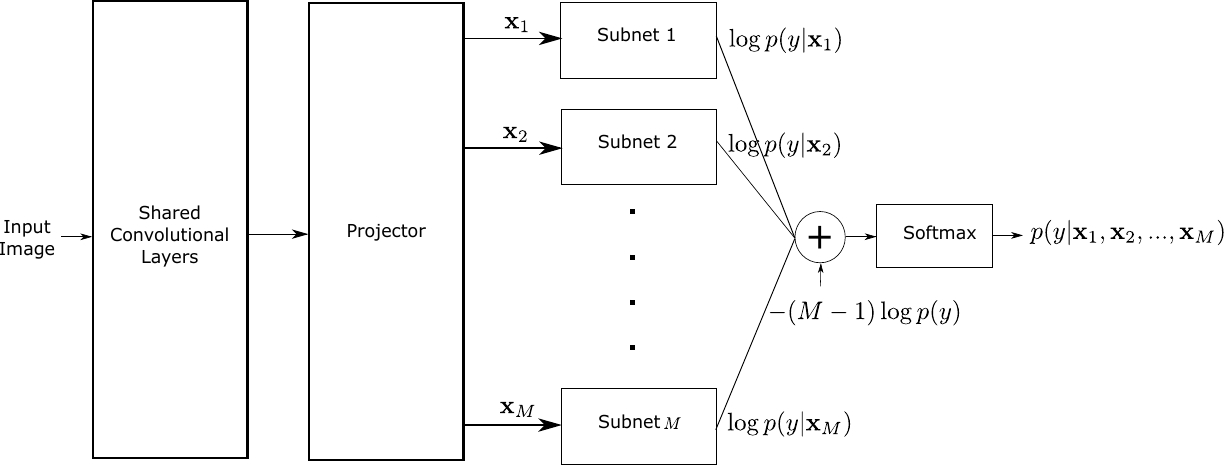}
	\caption{Superfeature-explaining teacher for CNNs.}
	\label{fig:cnn}
\end{figure}

An efficient approach for designing low-complexity CNN student models is to split the architecture along the filter dimension (see Figure \ref{fig:cnn}). Let $F$ be the number of output filters of the shared convolutional layers. If we split it into $M$ superfeatures, the input to each subnet will have only $F/M$ filters. We remark that the modern CNN architectures, particularly Resnets, are narrow and thus splitting into $M$ subnets often creates a bottleneck at each subnet for signal propagation. In such cases, we use a projector to increase the filter dimension before splitting into $M$ superfeatures. A projector $r:\mathbb{R}^{H\times W\times F}\rightarrow \mathbb{R}^{H\times W\times F'}$ is a function with $F'>F$, where $H\times W$ is the spatial dimension of the shared convolutional layer output. We use a 1$\times$1 convolutional layer as projector, which contains very few parameters. Such a linear projection also has whitening effect which decorrelates the filters and helps in promoting independence among the superfeatures. In Section \ref{sec:results}, we will see that with this approach, the KED type-$M$ teacher can achieve nearly identical classification accuracy as the original black-box CNN, leading to superior student performance.

\subsection{KED and Hidden-Representation Distillation Methods}
 
KED can be straightforwardly augmented with existing hidden-representation distillation methods. Let $h^g_m(\mathbf{x}_m)$, $1\leq m\leq M$, be the student's representations that we want to align with the teacher's intermediate layer outputs $h^f_m(\mathbf{x}_m)$, $1\leq m\leq M$. Adding the loss between the teacher's and the student's hidden representations, we rewrite (\ref{eq:kd_loss}) as
\begin{equation}
\begin{split}
\mathcal{L}(g)&=(1-\lambda)\mathbb{E}_\mathcal{D}[\mathcal{L}_\text{ce}(y, g(\mathbf{x}))]\\ 
&+ T^2\lambda(1-\mu)\mathbb{E}_\mathcal{D}[\mathcal{L}_\text{kl}(\sigma_T(f(\mathbf{x})), \sigma_T(g(\mathbf{x})))]\\
& + \frac{\tau^2\lambda\mu(1-\rho)}{M}\sum_{m=1}^{M}\mathbb{E}_\mathcal{D}[\mathcal{L}_\text{kl}(\sigma_\tau(f_m(\mathbf{x}_m)), \sigma_\tau(g_m(\mathbf{x}_m)))]\\ &+\frac{\lambda\mu\rho}{M}\sum_{m=1}^{M}\mathbb{E}_\mathcal{D}[\mathcal{L}_\text{h}(h^f_m(\mathbf{x}_m), h^g_m(\mathbf{x}_m))],
\end{split}
\label{eq:feat_loss}
\end{equation}
where $\rho$ is an additional hyperparameter. The loss term $\mathcal{L}_\text{h}$ is generic and can be set according to most existing hidden-representation distillation methods, e.g., FitNet \cite{Romero14}, attention transfer \cite{Zagoruyko17}, and similarity-preserving distillation \cite{Tung19}.

\subsection{Out-of-Distribution KED using Chimeric Set}

In scenarios where only a small training dataset is available, it is important to regularize the student. We propose to use the chimeric set as a natural technique for performance enhancement. For a type-$M$ teacher-student pair, let us rewrite the training dataset $\mathcal{X}$ as $\mathcal{X}=\{\mathcal{X}_1,...,\mathcal{X}_M\}$ where $\mathcal{X}_m$ denotes the set of m-th superfeatures among all training samples. Note that $|\mathcal{X}_m|=|\mathcal{X}|=K$. Then, we define the chimeric set as the $M$-fold Cartesian product $\mathcal{X}^M=\mathcal{X}_1\times...\times\mathcal{X}_M$. Thus, $|\mathcal{X}^M|=K^M$. This set contains only $K$ in-distribution samples and the rest are out-of-distribution (OOD) samples. 

Given a type-$M$ teacher's explanations for all the samples in the training dataset $\mathcal{X}$, we know the contribution of superfeatures in $\mathcal{X}_m, \forall m$. Therefore, we can calculate the teacher's explanations and predictions for all the samples in the chimeric set. Note that the chimeric samples do not have any true labels. However, to apply the framework of KED, we still treat the teacher's predictions converted to hard labels as if they were true labels. Then we can train a type-$M$ student over the chimeric set as described in Section \ref{subsec:ked}.

We remark that, for KD with a black-box teacher, training on the chimeric set may result in ``catastrophic forgetting'' \cite{McCloskey89} for black-box students, i.e., seeing OOD samples, the student may forget the information learned from the in-distribution samples. In contrast, for KED students, the chimeric set serves as a special data augmentation technique. This is different from other data augmentation method such as CutMix \cite{yun19}, particularly since here we take advantage of the architecture of type-$M$ students. When being trained on the chimeric set, a type-$M$ student always sees in-distribution superfeatures even in OOD samples. In other words, it learns from the same explanations but different predictions.

\section{Experimental Results}
\label{sec:results}

In this section, we study the performance of KED and compare it against that of conventional KD with black-box teachers. We also compare KED against KD when augmented with various hidden-representation distillation methods. Further experiments are conducted to evaluate the impact of the hyperparameters on KED and the performance of Algorithm \ref{algo:sf}.

\begin{table*}[t]
	\caption{Test accuracy for Unicauca with varying student's training dataset size.}
	\label{tab:unicauca}
	\centering
	\begin{tabular}{cccccc}
		\toprule
		\multirow{2}{*}{$|\mathcal{X}|$}	  & \multicolumn{2}{c}{Teacher}                      & \multicolumn{3}{c}{Student}\\ \cmidrule{2-6}
		& Black-box          & Type-$M$             & No Distl.     &  KD         &KED                          \\	\midrule
		10000  & \multirow{6}{*}{76.51\textpm0.04\%} & \multirow{6}{*}{77.33\textpm0.03\%} & 66.78\textpm0.04\% & 67.51\textpm0.04\% & \textbf{68.99\textpm0.04\%} \\
		30000         &  &  & 70.50\textpm0.04\% & 69.64\textpm0.04\% & \textbf{72.74\textpm0.04\%} \\
		50000        &  &  & 71.70\textpm0.04\% & 70.43\textpm0.04\% & \textbf{74.15\textpm0.04\%} \\
		70000        & &  & 72.88\textpm0.04\% & 71.47\textpm0.04\% & \textbf{74.76\textpm0.04\%} \\
		90000        & &  & 73.03\textpm0.04\% & 72.31\textpm0.04\% & \textbf{75.96\textpm0.03\%} \\
		108158 (Full)   & &  & 74.12\textpm0.04\% & 72.29\textpm0.04\% & \textbf{75.84\textpm0.04\%} \\
		\bottomrule
	\end{tabular}
\end{table*}

\begin{table*}[t]
	\caption{Test accuracy for MNIST with varying student's training dataset size.}
	\label{tab:mnist}
	\centering
	\begin{tabular}{cccccc}
		\toprule
		\multirow{2}{*}{$|\mathcal{X}|$}	  & \multicolumn{2}{c}{Teacher}                      & \multicolumn{3}{c}{Student}\\ \cmidrule{2-6}
		& Black-box          & Type-$M$             & No Distl.     &  KD         &KED                          \\	\midrule
		10000  & \multirow{6}{*}{98.43\textpm0.02\%} & \multirow{6}{*}{98.40\textpm0.02\%} & 93.32\textpm0.03\% & 95.15\textpm0.03\% & \textbf{96.87\textpm0.02\%} \\
		20000         &  &  & 94.31\textpm0.04\% & 95.64\textpm0.03\% & \textbf{97.28\textpm0.02\%} \\
		30000        &  &  & 94.81\textpm0.03\% & 96.33\textpm0.03\% & \textbf{97.32\textpm0.02\%} \\
		40000        & &  & 95.32\textpm0.03\% & 96.37\textpm0.03\% & \textbf{97.44\textpm0.02\%} \\
		50000        & &  & 95.49\textpm0.03\% & 96.78\textpm0.03\% & \textbf{97.63\textpm0.02\%} \\
		60000 (Full)   & &  & 95.41\textpm0.03\% & 96.69\textpm0.02\% & \textbf{97.52\textpm0.02\%} \\
		\bottomrule
	\end{tabular}
\end{table*}

\begin{table*}[t]
	\caption{Test accuracy for FashionMNIST with varying student's training dataset size.}
	\label{tab:fmnist}
	\centering
	\begin{tabular}{cccccc}
		\toprule
		\multirow{2}{*}{$|\mathcal{X}|$}	  & \multicolumn{2}{c}{Teacher}                      & \multicolumn{3}{c}{Student}\\ \cmidrule{2-6}
		& Black-box          & Type-$M$             & No Distl.     &  KD         &KED                          \\	\midrule
		10000  & \multirow{6}{*}{89.98\textpm0.04\%} & \multirow{6}{*}{90.16\textpm0.04\%} & 84.86\textpm0.05\% & 85.31\textpm0.05\% & \textbf{87.50\textpm0.05\%} \\
		20000         &  &  & 85.98\textpm0.04\% & 86.78\textpm0.04\% & \textbf{88.58\textpm0.04\%} \\
		30000        &  &  & 86.69\textpm0.04\% & 87.38\textpm0.04\% & \textbf{88.41\textpm0.05\%} \\
		40000        & &  & 87.18\textpm0.05\% & 87.49\textpm0.05\% & \textbf{88.79\textpm0.04\%} \\
		50000        & &  & 87.88\textpm0.05\% & 87.47\textpm0.05\% & \textbf{88.88\textpm0.05\%} \\
		60000 (Full)   & &  & 87.96\textpm0.05\% & 88.11\textpm0.05\% & \textbf{89.38\textpm0.04\%} \\
		\bottomrule
	\end{tabular}
\end{table*}

\begin{table*}[ht]
	\caption{Test accuracy for CIFAR10 with varying model architectures.}
	\label{tab:cifar10}
	\centering
	\begin{tabular}{lllllllll}
		\toprule
		Teacher    & Resnet44x2                        & Resnet56x2 & Resnet56x2 & WRN-16-8                        & WRN-16-8          & WRN-28-4                & WRN-28-4                          & VGG13                        \\ \midrule
		Black-box  & 94.47\textpm0.03\% & 94.22\textpm0.03\% &94.22\textpm 0.03\% & 95.02\textpm0.03\% & 95.02\textpm0.03\% & 94.81\textpm0.03\% & 94.81\textpm0.03\% & 93.35\textpm0.04\% \\
		Type-$M$   & 94.37\textpm0.03\% & 94.37\textpm0.03\% & 94.37\textpm0.03\% & 94.75\textpm0.03\% & 94.75\textpm0.03\% & 94.95\textpm0.03\% & 94.95\textpm0.03\% & 92.90\textpm0.04\% \\ \midrule
		Student    & WRN-10-1                          & Resnet8  & WRN-10-1  & Resnet8               & WRN-16-1    & Resnet8                      & VGG8                              & VGG8                              \\ \midrule
		No Distl.  & 88.44\textpm0.05\% & 87.98\textpm0.05\% & 88.44\textpm0.05\% & 87.98\textpm0.05\% & 91.42\textpm0.04\% & 87.98\textpm0.05\% & 90.93\textpm0.04\% & 90.93\textpm0.04\% \\
		KD         & 88.69\textpm0.04\% & 88.53\textpm0.04\% &88.39\textpm0.05\% &88.96\textpm0.04\% & 91.57\textpm0.04\% &88.68\textpm0.04\% & 90.26\textpm0.05\% & 90.86\textpm0.04\% \\
		KED        & 89.88\textpm0.04\% & \textbf{89.43\textpm0.04\%} &\textbf{89.82\textpm0.04\%} &\textbf{89.70\textpm0.05\%} & 91.89\textpm0.04\% & \textbf{89.38\textpm0.04\%} & 92.13\textpm0.04\% & 91.57\textpm0.04\% \\
		KD+FitNet  & 89.17\textpm0.04\% & 88.67\textpm0.05\% & 89.42\textpm0.04\% &88.94\textpm0.04\% & 91.83\textpm0.04\% &88.98\textpm0.04\% & 91.31\textpm0.04\% & 91.07\textpm0.04\% \\
		KED+FitNet & \textbf{90.08\textpm0.04\%} & 89.06\textpm0.05\% &89.62\textpm0.04\% &89.36\textpm0.04\% & 92.04\textpm0.04\% &89.19\textpm0.05\% & 91.77\textpm0.04\% & 91.87\textpm0.04\% \\
		KD+AT      & 88.98\textpm0.04\% & 88.32\textpm0.05\% &89.04\textpm0.05\% &88.26\textpm0.05\% & 92.07\textpm0.04\% &88.15\textpm0.05\% & 90.96\textpm0.04\% & 91.29\textpm0.04\% \\
		KED+AT     & 90.03\textpm0.04\% & 89.21\textpm0.05\% &89.31\textpm0.04\% &89.30\textpm0.04\% &92.18\textpm0.04\% & 89.05\textpm0.04\% & 91.81\textpm0.04\% & 91.69\textpm0.04\% \\
		KD+SP      & 88.92\textpm0.04\% & 88.43\textpm0.04\% &89.10\textpm0.04\% &88.87\textpm0.05\% & 92.18\textpm0.04\% &88.64\textpm0.04\% & 91.79\textpm0.04\% & 91.15\textpm0.04\% \\
		KED+SP     & 89.43\textpm0.05\% & 88.95\textpm0.04\% &89.67\textpm0.04\% &89.54\textpm0.05\% & \textbf{92.39\textpm0.03\%} &89.09\textpm0.04\% & \textbf{92.36\textpm0.04\%} & \textbf{91.94\textpm0.04\%} \\
		\bottomrule
	\end{tabular}
\end{table*}

\begin{table*}[ht]
	\caption{Test accuracy for CIFAR100 with varying model architectures.}
	\label{tab:cifar100}
	\centering
	\begin{tabular}{lllllllll}
		\toprule
		Teacher    & Resnet44x2                        & Resnet56x2  & Resnet56x2      &WRN-16-8               & WRN-16-8                       & WRN-28-4   & WRN-28-4                          & VGG13                             \\ \midrule
		Black-box  & 75.11\textpm0.06\% & 75.48\textpm0.06\% & 75.48\textpm0.06\% & 77.89\textpm0.05\% & 77.89\textpm0.05\% &76.73\textpm0.06\% & 76.73\textpm0.06\% & 72.98\textpm0.06\% \\
		Type-$M$   & 74.96\textpm0.06\% & 75.17\textpm0.06\% &75.17\textpm0.06\% & 78.17\textpm0.05\% & 78.17\textpm0.05\% & 76.98\textpm0.06\% & 76.98\textpm0.06\% & 71.86\textpm0.06\% \\ \midrule
		Student    & WRN-10-2             & Resnet20        & WRN-10-2  & Resnet20            & WRN-16-2     & Resnet20                     & VGG8                              & VGG8                              \\ \midrule
		No Distl.  & 69.01\textpm0.06\% & 68.30\textpm0.07\% & 69.01\textpm0.06\% & 68.30\textpm0.07\% & 72.25\textpm0.06\% & 68.30\textpm0.07\% & 68.72\textpm0.06\% & 68.72\textpm0.06\% \\
		KD         & 69.47\textpm0.06\% & 69.19\textpm0.06\% & 69.59\textpm0.06\% & 69.29\textpm0.07\% & 74.34\textpm0.06\% & 69.66\textpm0.06\% & 70.39\textpm0.06\% & 69.95\textpm0.06\% \\
		KED        & 71.33\textpm0.06\% & 70.94\textpm0.06\% & 71.12\textpm0.07\% &70.89\textpm0.06\% & 74.94\textpm0.06\% &\textbf{71.36\textpm0.06\%} & \textbf{73.50\textpm0.06\%} & 71.30\textpm0.06\% \\
		KD+FitNet  & 69.91\textpm0.06\% & 69.42\textpm0.07\% & 69.98\textpm0.07\% & 69.36\textpm0.06\% & 74.37\textpm0.06\% &69.60\textpm0.06\% & 70.89\textpm0.07\% & 69.98\textpm0.06\% \\
		KED+FitNet & 71.36\textpm0.06\% & \textbf{71.79\textpm0.07\%} & \textbf{71.28\textpm0.06\%} & \textbf{71.39\textpm0.06\%} & \textbf{75.09\textpm0.06\%} &71.18\textpm0.06\% & 72.94\textpm0.06\% & 70.92\textpm0.06\% \\
		KD+AT      & 69.53\textpm0.06\% & 69.24\textpm0.06\% &69.50\textpm0.06\% &69.04\textpm0.06\% & 73.82\textpm0.06\% &69.02\textpm0.06\% & 70.29\textpm0.07\% & 69.27\textpm0.07\% \\
		KED+AT     & 70.84\textpm0.07\% & 70.38\textpm0.06\% &70.57\textpm0.06\% & 70.98\textpm0.06\% & 74.57\textpm0.06\% & 70.75\textpm0.06\% & 72.61\textpm0.06\% & 70.84\textpm0.06\% \\
		KD+SP      & 69.83\textpm0.06\% & 69.57\textpm0.06\% &69.35\textpm0.06\% & 68.69\textpm0.07\% & 73.68\textpm0.06\% & 68.79\textpm0.06\% & 69.87\textpm0.06\% & 70.40\textpm0.06\% \\
		KED+SP     & \textbf{71.42\textpm0.06\%} & 70.81\textpm0.06\% &71.23\textpm0.06\% & 70.52\textpm0.06\% & 74.67\textpm0.06\% & 71.19\textpm0.06\% & 72.07\textpm0.06\% & \textbf{71.67\textpm0.06\%} \\ 
		\bottomrule
	\end{tabular}
\end{table*}

\begin{table*}[ht]
	\caption{Test accuracy for Tiny Imagenet with varying model architectures.}
	\label{tab:tinyimg}
	\centering
	\begin{tabular}{lllllllll}
		\toprule
		Teacher    & Resnet44x2                        & Resnet56x2   & Resnet56x2                      & WRN-16-8   & WRN-16-8           & WRN-28-4              & WRN-28-4                          & VGG13                             \\ \midrule
		Black-box  & 61.04\textpm0.07\% & 61.09\textpm0.07\% & 61.09\textpm0.07\% &63.46\textpm0.07\% & 63.46\textpm0.07\% &62.60\textpm0.07\% & 62.60\textpm0.07\% & 59.32\textpm0.07\% \\
		Type-$M$   & 60.79\textpm0.07\% & 60.80\textpm0.07\% &60.80\textpm0.07\% &63.35\textpm0.07\% & 63.35\textpm0.07\% &61.99\textpm0.07\% & 61.99\textpm0.07\% & 58.88\textpm0.07\% \\ \midrule
		Student    & WRN-10-2                          & Resnet20    & WRN-10-2                          & Resnet20                      & WRN-16-2        & Resnet20                  & VGG8                              & VGG8                              \\ \midrule
		No Distl.  & 49.39\textpm0.08\% & 51.84\textpm0.07\% &49.39\textpm0.08\% &51.84\textpm0.07\% & 56.13\textpm0.07\% &51.84\textpm0.07\% & 53.50\textpm0.07\% & 53.50\textpm0.07\% \\
		KD         & 50.52\textpm0.07\% & 52.72\textpm0.07\% &51.26\textpm0.07\% &52.88\textpm0.07\% & 57.38\textpm0.07\% &52.47\textpm0.07\% & 57.55\textpm0.07\% & 56.34\textpm0.06\% \\
		KED        & 52.98\textpm0.07\% & 54.96\textpm0.07\% &53.18\textpm0.07\% &53.87\textpm0.07\% & 58.72\textpm0.07\% &54.94\textpm0.07\% & 60.42\textpm0.07\% & 59.06\textpm0.07\% \\
		KD+FitNet  & 53.01\textpm0.07\% & 53.71\textpm0.07\% &53.35\textpm0.07\% &52.97\textpm0.07\% & 59.41\textpm0.07\% &53.27\textpm0.07\% & 58.55\textpm0.07\% & 56.28\textpm0.07\% \\
		KED+FitNet & \textbf{54.79\textpm0.07\%} & 54.90\textpm0.07\% & \textbf{55.27\textpm0.07\%} &54.43\textpm0.07\% & \textbf{60.23\textpm0.07\%} &55.40\textpm0.07\% & 60.74\textpm0.07\% & 58.69\textpm0.07\% \\
		KD+AT      & 51.96\textpm0.07\% & 53.32\textpm0.07\% &52.33\textpm0.07\% &52.22\textpm0.07\% & 59.14\textpm0.07\% &53.04\textpm0.07\% & 57.79\textpm0.07\% & 54.63\textpm0.07\%   \\
		KED+AT     & 53.82\textpm0.07\% & \textbf{55.35\textpm0.08\%} &54.02\textpm0.07\% &\textbf{54.59\textpm0.07\%} & 59.93\textpm0.07\% & \textbf{55.10\textpm0.07\%} & \textbf{60.99\textpm0.07\%} & 58.59\textpm0.07\% \\
		KD+SP      & 52.08\textpm0.07\% & 52.59\textpm0.07\% &52.25\textpm0.07\% &52.49\textpm0.07\% & 58.44\textpm0.07\% &52.75\textpm0.07\% & 57.35\textpm0.07\% & 58.52\textpm0.07\% \\
		KED+SP     & 54.46\textpm0.07\% & 54.75\textpm0.08\% & 54.36\textpm0.08\% & 53.59\textpm0.07\% &  59.70\textpm0.07\% & 54.38\textpm0.07\% & 60.14\textpm0.07\% & \textbf{59.47\textpm0.07\%} \\
		\bottomrule
	\end{tabular}
\end{table*}

\subsection{Datasets and Experimental Setup}
\label{subsec:setup}

We run experiments on classification over various datasets, including the Unicauca network traffic dataset \cite{unicauca_url}, MNIST \cite{mnist}, FashionMNIST \cite{fmnist}, CIFAR10 \cite{cifar10}, CIFAR100 \cite{cifar10}, and Tiny Imagenet \cite{Tinyimg}. We implement multi-layer perceptrons (MLPs), Resnets \cite{He16}, wide Resnets \cite{Zagoruyko16}, and VGGs for the teacher and student models.

For the Unicauca dataset, we group the network traffic application types into three delay classes and categorize flows according to their delay sensitivity. In KD, the teacher has size [200, 200, 200, 200], i.e., four hidden layers with 200 neurons each. The architecture of the student is [50, 50, 50, 50]. For MNIST and FashionMNIST, the KD teacher has size [500, 500]. The black-box student in KD has size [20, 20] for MNIST and [60, 60] for FashionMNIST. For KED, we construct the superfeature-explaining teacher and student by choosing the number of neurons per hidden layer such that they have a similar number of total model parameters as the benchmark KD with a black-box teacher. In particular, we note that a type-$M$ model with $L$ hidden layers has $M(L-1)n_h^2+ (ML+MC+d)n_h+MC$ parameters where $n_h$ is the number of neurons per hidden layer. Thus, we set this expression equal to the number of model parameters in black-box KD and solve for $n_h$.

For CIFAR10, CIFAR100 and Tiny Imagenet, we use standard Resnets, wide Resnets and VGGs. We use the labels ``Resnet$n$x$k$,'' ``WRN-$n$-$k$,'' and ``VGG$n$'' where the variables $n$ and $k$ denote the number of layers and the widening factor, respectively. For KED, type-$M$ models of similar complexity are constructed by splitting the last stack of convolutional and fully connected layers in the aforementioned models.

For all experiments, the default setting for the hyperparameters is $T=10$, $\tau=10$, $\lambda=0.7$, $\mu=0.7$, and $\rho=0.7$. We keep the number of superfeatures fixed at $M=4$. For the Unicauca, MNIST, and FashionMNIST datasets, the teachers and the students, respectively, are trained for 100 epochs with a batch size of 500 and 100. We use RELU activation and the Adam optimizer with a learning rate of $0.001$. The prior for type-$M$ models $p(y)$ is estimated by taking a sample average of the black-box teacher's predictions over the training set. For picking superfeatures, we apply Algorithm \ref{algo:sf} computing the average Hessian of the black-box teacher over a random subset of 1000 training samples. We use the Louvain community detection method in scikit-network version 0.27.1 \cite{sknetwork}. We tune the resolution with a stepsize of 0.01. For CIFARs and Tiny Imagenet, we train the teachers and the students for 150 and 75 epochs, respectively, with a batch size of 100. For data augmentation, we pad 4 pixels on each side of a training image, and then apply random crop and random horizontal flip to it. We use RELU activation and the SGD optimizer with Nesterov momentum 0.9. The initial learning rate is set to 0.01 for the first 10 epochs as warm-up period. Then, for CIFARs, we use a learning rate of 0.05 and divide it by 10 after 120 and 140 epochs. The L2 regularization coefficient is set to $5\times10^{-4}$. For Tiny Imagenet, we use a learning rate of 0.1 and divide it by 10 after 60 and 70 epochs. The L2 regularization coefficient is set to $1\times10^{-4}$. As described in Section \ref{subsec:cnns}, for CNNs, the superfeatures are created from the output of a projector, which is a 1x1 convolutional layer. We assume uniform prior over the classes and thus eliminate $p(y)$ from (\ref{eq:sf}). 

For the experiments on the chimeric set, we randomly generate one million samples and continue training the distilled students for 5 epochs with a batch size of 100. To avoid numerical instability, we have added a small bias of $10^{-15}$ to the argument of the $\log$ function. In all experiments, we obtain a 95\% confidence intervals for inference by bootstrapping the test set.

\subsection{Distillation Performance with MLPs}

Table \ref{tab:unicauca} presents the test accuracy of KED for the Unicauca dataset varying the student's training dataset size. We observe that KED substantially outperforms KD and no distillation. For example, the student with 10000 samples in the training dataset achieves 66.78\% and 67.51\% accuracy without and with KD, respectively. The KED student achieves an accuracy of 68.99\%. On the full dataset, the black-box student achieves 74.12\% and 72.29\% accuracy without and with KD, respectively, whereas the KED student reaches an accuracy of 75.84\%. In this case, KD fails to improve the performance of the student. In contrast, KED still benefits the student significantly. 

We show further experimental results for the MNIST and FashionMNIST datasets in Tables \ref{tab:mnist} and \ref{tab:fmnist}, respectively. For MNIST, with 10000 samples in the training dataset, the black-box student achieves an accuracy of 93.32\%, which increases to 95.15\% under KD. In comparison, the KED student achieves an improved accuracy of 96.87\%. When trained on the full dataset, the black-box student provides 95.41\% and 96.69\% accuracy without and with KD, respectively. The KED student reaches an accuracy of 97.52\%.

We see a similar trend for the FashionMNIST dataset as well. Training the black-box student using 10000 samples results in a test accuracy of 84.86\% and 85.31\% under no distillation and KD, respectively. In comparison, the KED student attains 87.50\% accuracy. Furthermore, when trained on the full dataset, the black-box student provides an accuracy of 87.96\% and 88.11\% without and with KD, respectively. KED enables the student to achieve 89.38\% accuracy.

In all of these experiments, we observe that the accuracy gap between the KED teacher and the KED student is much narrower compared with their black-box KD counterparts. This clearly demonstrates that a superfeature-explaining teacher can transfer more knowledge than a black-box teacher of similar complexity.

\subsection{Distillation Performance with CNNs}

In Tables \ref{tab:cifar10}, \ref{tab:cifar100} and \ref{tab:tinyimg}, we present the test accuracy of KED with CNNs over full datasets of CIFAR10, CIFAR100 and Tiny Imagenet, respectively. Here the labels FitNet, AT, and SP indicate the hidden-representation methods of \cite{Romero14}, \cite{Zagoruyko17}, and \cite{Tung19}, respectively.

Let us consider Resnet56x2 over CIFAR10 as shown in Table \ref{tab:cifar10}. When distilled to Resnet8 using KD, we achieve 88.53\% accuracy. However, the KED student outperforms the KD student, reaching 89.43\% accuracy. We notice a similar trend for all other architectures.

Furthermore, take Resnet56x2 over CIFAR100 for another example, which is shown in Table \ref{tab:cifar100}. When distilled to Resnet20 using KD, we achieve 69.19\% accuracy. However, the KED student outperforms KD reaching 70.94\% accuracy. Augmented with FitNet, the KED student further improves to 71.79\%, but the same augmentation of KD reaches only 69.42\%. We observe a similar trend for all other combinations of teacher and student architectures.

As shown in Table \ref{tab:tinyimg}, for Tiny Imagenet, when the teacher is a Resnet56x2 model, a black-box Resnet20 student achieves an accuracy of 52.72\% with KD. In KED, the student attains 54.96\% test accuracy. Augmenting KED with attention transfer, the student's performance further improves to 55.35\%, while the same augmentation of KD achieves only 53.71\%. More importantly, the gap between the KED teacher and the KED student is always significantly narrower than in KD.

\subsection{Benefits of Chimeric Set}
\label{app:chimeric}

We show the performance of knowledge distillation using the chimeric set in Tables \ref{tab:chi_unicauca}, \ref{tab:chi_mnist}, and \ref{tab:chi_fmnist}. We run these experiments with fewer than 10000 samples in the student's training dataset to illustrate the impact of the chimeric set in a limited-data scenario. For each of these experiments, the student is first trained via the regular KD or KED and then we continue training it on the chimeric set. We observe that for many cases in MNIST and FashionMNIST, the chimeric set improves the performance of both KD and KED. However, for Unicauca, only KED benefits significantly from the chimeric set.

\begin{table*}[ht]
	\caption{Test accuracy for Unicauca using chimeric set while varying the student's training dataset size.}
	\centering
	\begin{tabular}{cccccc}
		\toprule
		Student's training dataset size & KED & KED with chimeric set & KD & KD with chimeric set & No distillation \\ \midrule
		2000                            & 63.61\% \textpm 0.04\% & 65.51\% \textpm 0.04\% & 60.56\% \textpm 0.04\% & 59.03\% \textpm 0.04\% & 59.95\% \textpm 0.04\%     \\ 
		4000                            & 66.63\% \textpm 0.04\% & 68.74\% \textpm 0.04\% & 62.92\% \textpm 0.04\% & 59.90\% \textpm 0.04\% & 62.83\% \textpm 0.04\%             \\ 
		6000                            & 68.07\% \textpm 0.04\% & 69.79\% \textpm 0.04\% & 65.49\% \textpm 0.04\% & 61.31\% \textpm 0.04\% & 64.55\% \textpm 0.04\%             \\ 
		8000                            & 69.37\% \textpm 0.04\% & 71.10\% \textpm 0.04\% & 66.22\% \textpm 0.04\% & 61.19\% \textpm 0.04\% & 66.42\% \textpm 0.04\%             \\ \bottomrule
	\end{tabular}
	\label{tab:chi_unicauca}
\end{table*}

\begin{table*}[ht]
	\caption{Test accuracy for MNIST using chimeric set while varying the student's training dataset size.}
	\centering
	\begin{tabular}{cccccc}
		\toprule
		Student's training dataset size & KED & KED with chimeric set & KD & KD with chimeric set & No distillation \\ \midrule
		2000                            & 95.23\% \textpm 0.03\% & 96.26\% \textpm 0.02\% & 91.38\% \textpm 0.04\% & 92.69\% \textpm 0.04\%  & 90.27\% \textpm 0.04\%     \\ 
		4000                            & 96.01\% \textpm 0.03\% & 96.92\% \textpm 0.02\% & 93.89\% \textpm 0.04\% & 93.07\% \textpm 0.03\% & 91.72\% \textpm 0.04\%             \\ 
		6000                            & 96.57\% \textpm 0.02\% & 96.92\% \textpm 0.02\% & 94.41\% \textpm 0.03\% & 93.24\% \textpm 0.04\% & 92.63\% \textpm 0.04\%             \\ 
		8000                            & 96.91\% \textpm 0.02\% & 96.98\% \textpm 0.02\% & 94.62\% \textpm 0.04\% & 90.44\% \textpm 0.04\% & 93.15\% \textpm 0.04\%         \\	\bottomrule
	\end{tabular}
	\label{tab:chi_mnist}
\end{table*}

\begin{table*}[ht]
	\caption{Test accuracy for FashionMNIST using chimeric set while varying the student's training dataset size.}
	\centering
	\begin{tabular}{cccccc}
		\toprule
		Student's training dataset size & KED & KED with chimeric set & KD & KD with chimeric set & No distillation \\ \midrule
		2000                            & 84.89\% \textpm 0.05\% & 85.44\% \textpm 0.05\% & 82.52\% \textpm 0.06\% & 84.60\% \textpm 0.05\% & 81.14\% \textpm 0.05\%  \\
		4000                            & 86.26\% \textpm 0.05\% & 87.27\% \textpm 0.05\% & 83.32\% \textpm 0.05\% & 84.77\% \textpm 0.05\% & 83.58\% \textpm 0.06\%             \\ 
		6000                            & 87.46\% \textpm 0.05\% & 87.91\% \textpm 0.05\% & 84.45\% \textpm 0.05\% & 84.53\% \textpm 0.05\% & 84.17\% \textpm 0.06\%            \\
		8000                            & 87.51\% \textpm 0.05\% & 88.16\% \textpm 0.04\% & 84.86\% \textpm 0.05\% & 84.76\% \textpm 0.05\% & 84.53\% \textpm 0.06\%             \\ \bottomrule
	\end{tabular}
	\label{tab:chi_fmnist}
\end{table*}

\subsection{Impact of Hyperparameter Values and Ablation Studies}
\label{app:hyper}

We conduct further experiments to study the impact of the hyperparameters $M$, $T$, $\tau$, $\lambda$ and $\mu$ on KED and to evaluate the performance of Algorithm \ref{algo:sf}. We perform these experiments on Unicauca and MNIST with 10000 samples in the student's training dataset. We also provide ablation studies on CIFAR10, CIFAR100, and Tiny Imagenet.

\subsubsection{Impact of the Number of Superfeatures, $M$}

We demonstrate the impact of the number of superfeatures on KED. In Table \ref{tab:sf_M}, we show type-$M$ teacher's and type-$M$ student's test accuracy on Unicauca and MNIST varying the number of superfeatures $M$. Increasing $M$ has two opposing effects on the student's performance. On one hand, a larger $M$ implies more explanation from the teacher. On the other hand, both the teacher's and the student's architecture becomes more restricted with respect to modeling the dependency among the superfeatures. In particular, we observe some degradation in the teacher’s performance as $M$ increases. Thus, the student's performance improves with increasing $M$ only when $M$ is not too large.

\begin{table*}[ht]
	\caption{Test accuracy of KED on Unicauca and MNIST, varying the number of superfeatures $M$.}
	\label{tab:sf_M}
	\centering
	\begin{tabular}{ccccc}
		\toprule
		\multirow{2}{*}{\# Superfeatures} & \multicolumn{2}{c}{Unicauca}    & \multicolumn{2}{c}{MNIST}    \\ \cmidrule{2-5} 
		& \multicolumn{1}{c}{Teacher} & Student  & \multicolumn{1}{c}{Teacher} & Student \\ \midrule
		$M=2$	& 76.80\textpm0.04\% & 68.15\textpm0.04\% & 98.31\textpm0.02\% & 96.49\textpm0.02\% \\ 
		$M=4$	& 77.33\textpm0.03\% & 68.99\textpm0.04\% & 98.40\textpm0.02\% & 96.87\textpm0.02\%  \\
		$M=8$	& 76.71\textpm0.03\% & 70.13\textpm0.04\% & 98.18\textpm0.02\% & 96.67\textpm0.03\%  \\
		$M=16$	& 76.23\textpm0.04\% & 70.09\textpm0.04\% & 97.94\textpm0.02\% & 96.65\textpm0.02\%  \\ 
		\bottomrule
	\end{tabular}
\end{table*}

\subsubsection{Impact of Temperature Parameters, $T$ and $\tau$}

In Tables \ref{tab:temp_unicauca} and \ref{tab:temp_mnist}, we present the impact of temperature parameters $T$ and $\tau$ on Unicauca and MNIST, respectively. For Unicauca, the accuracy varies between 68.34\% at $T=20$ and $\tau=10$, and 70.45\% at $T=1$ and $\tau=1$. For MNIST, the minimum accuracy of 96.21\% is observed at $T=20$ and $\tau=1$, and the maximum accuracy of 97.21\% is achieved at $T=10$ and $\tau=20$. We note that the choice of distillation temperature significantly affects the performance in both cases.

\begin{table*}[ht]
	\caption{Test accuracy of KED student on Unicauca, varying the temperature parameters, $T$ and $\tau$.}
	\label{tab:temp_unicauca}
	\centering
	\begin{tabular}{ccccc}
		\toprule
		Temperature  				& $T=1$ & $T=5$ & $T=10$ & $T=20$ \\ \midrule
		$\tau=1$                    & 70.45\textpm0.04\% & 69.65\textpm0.04\% & 68.56\textpm0.04\% & 68.86\textpm0.04\%      \\ 
		$\tau=5$                    & 70.14\textpm0.04\% & 69.24\textpm0.04\% & 69.34\textpm0.04\% & 68.52\textpm0.04\%       \\
		$\tau=10$                   & 70.09\textpm0.04\%  & 69.39\textpm0.04\% & 68.99\textpm0.04\% & 68.34\textpm0.04\%          \\
		$\tau=20$                   & 69.77\textpm0.04\% & 69.49\textpm0.04\% & 68.82\textpm0.04\% & 68.77\textpm0.04\%          \\
		\bottomrule
	\end{tabular}
\end{table*}

\begin{table*}[ht]
	\caption{Test accuracy of KED student on MNIST, varying the temperature parameters, $T$ and $\tau$.}
	\label{tab:temp_mnist}
	\centering
	\begin{tabular}{ccccc}
		\toprule
		Temperature  				& $T=1$ & $T=5$ & $T=10$ & $T=20$ \\ \midrule
		$\tau=1$                    & 96.59\textpm0.03\% & 96.37\textpm0.03\% & 96.49\textpm0.02\% & 96.21\textpm0.02\%      \\
		$\tau=5$                    & 96.61\textpm0.02\% & 96.71\textpm0.02\% & 96.95\textpm0.02\% & 96.39\textpm0.02\%       \\
		$\tau=10$                   & 96.85\textpm0.02\%  & 96.92\textpm0.02\% & 96.87\textpm0.02\% & 96.68\textpm0.02\%          \\
		$\tau=20$                   & 96.72\textpm0.03\%  & 97.02\textpm0.02\% & 97.21\textpm0.02\% & 96.75\textpm0.02\%          \\
		\bottomrule
	\end{tabular}
\end{table*}

\subsubsection{Impact of Loss Weights, $\lambda$ and $\mu$}

We observe the impact of loss weights $\lambda$ and $\mu$ in Tables \ref{tab:weight_unicauca} and \ref{tab:weight_mnist}, on Unicauca and MNIST, respectively. For Unicauca, the minimum accuracy of 67.90\% is observed at $\lambda=0.8$ and $\mu=0.2$, and the maximum accuracy of 69.66\% is achieved at $\lambda=0.2$ and $\mu=0.6$. For MNIST, the accuracy varies between 96.39\% at $\lambda=0.6$ and $\mu=0.2$, and 97.03\% at $\lambda=0.6$ and $\mu=0.8$. We note that the choice of loss weights significantly affects the performance in Unicauca but not as much in MNIST.     

\begin{table*}[ht]
	\caption{Test accuracy of KED student on Unicauca, varying the loss weights, $\lambda$ and $\mu$.}
	\label{tab:weight_unicauca}
	\centering
	\begin{tabular}{ccccc}
		\toprule
		Loss weights  				& $\lambda=0.2$ & $\lambda=0.4$ & $\lambda=0.6$ & $\lambda=0.8$ \\ \midrule
		$\mu=0.2$                   & 69.29\textpm0.04\% & 68.83\textpm0.04\% & 68.41\textpm0.04\% & 67.90\textpm 0.04\%      \\ 
		$\mu=0.4$                   & 69.09\textpm0.04\% & 68.81\textpm0.04\% & 68.75\textpm0.04\% & 68.53\textpm 0.04\%       \\ 
		$\mu=0.6$                   & 69.66\textpm0.04\%  & 68.96\textpm0.04\% & 69.35\textpm0.04\% & 68.55\textpm0.04\%          \\ 
		$\mu=0.8$                   & 69.41\textpm0.04\%  & 69.33\textpm0.04\% & 69.05\textpm0.04\% & 68.96\textpm0.04\%          \\ 
		\bottomrule
	\end{tabular}
\end{table*}

\begin{table*}[ht]
	\caption{Test accuracy of KED student on MNIST, varying the loss weights, $\lambda$ and $\mu$.}
	\label{tab:weight_mnist}
	\centering
	\begin{tabular}{ccccc}
		\toprule
		Loss weights  				& $\lambda=0.2$ & $\lambda=0.4$ & $\lambda=0.6$ & $\lambda=0.8$ \\ \midrule
		$\mu=0.2$                   & 96.40\textpm0.02\% & 96.56\textpm0.03\% & 96.39\textpm0.03\% & 96.43\textpm0.03\%      \\ 
		$\mu=0.4$                   & 96.49\textpm0.03\% & 96.79\textpm0.02\% & 96.72\textpm0.03\% & 96.83\textpm0.03\%       \\ 
		$\mu=0.6$                   & 96.65\textpm0.03\%  & 96.83\textpm0.02\% & 96.67\textpm0.03\% & 96.83\textpm0.02\%          \\ 
		$\mu=0.8$                   & 96.89\textpm0.02\%  & 96.74\textpm0.02\% & 97.03\textpm0.02\% & 96.92\textpm0.02\%          \\ \bottomrule
	\end{tabular}
\end{table*}

\subsubsection{Ablation Studies}

In Table \ref{tab:ablation}, we show the impact of the KED teacher's prediction and explanation on the performance of the KED student. Note that even in the absence of the teacher's explanation, the KED student can learn the contribution of superfeatures just from the teacher's prediction. This is because of the superfeature-explaining architecture of both the teacher and the student. However, only when teacher's prediction and explanation are provided simultaneously, the student achieves the best performance.  

\begin{table*}[ht]
	\caption{Ablation studies with KED student (Teacher: WRN-28-4, Student: Resnet8 / Resnet20) }
	\label{tab:ablation}
	\centering
	\begin{tabular}{cccccc}
		\toprule
		Loss weights & \shortstack{Teacher's\\Prediction} & \shortstack{Teacher's\\Explanation} & CIFAR10 & CIFAR100 & Tiny Imagenet \\ \midrule
		$\lambda=0.7, \mu=0.0$   & \cmark & \xmark & 89.15\textpm0.04\% & 70.57\textpm0.06\%  & 54.82\textpm0.06\%   \\ 
		$\lambda=0.7, \mu=1.0$   & \xmark & \cmark & 88.97\textpm0.04\% & 69.35\textpm0.06\%   &   53.43\textpm0.07\% \\ 
		$\lambda=0.7, \mu=0.7$ & \cmark  & \cmark & 89.38\textpm0.04\% & 71.36\textpm0.06\%     &  54.94\textpm0.07\%  \\ 
		\bottomrule
	\end{tabular}
\end{table*}

\subsection{Evaluation of Algorithm \ref{algo:sf} on MNIST+FashionMNIST Combined Dataset}
\label{app:algo}

\begin{figure}[ht]
	\centering
	\begin{subfigure}{.33\linewidth}
		\centering
		\includegraphics[width=2.5cm]{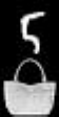}
		\caption{Image}
		\label{fig:img}
	\end{subfigure}%
	\begin{subfigure}{.3\linewidth}
		\centering
		\includegraphics[width=2.5cm]{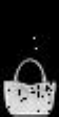}
		\caption{Superfeature 1}
		\label{fig:sf_1}
	\end{subfigure}
	\begin{subfigure}{.3\linewidth}
		\centering
		\includegraphics[width=2.5cm]{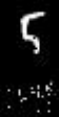}
		\caption{Superfeature 2}
		\label{fig:sf_2}
	\end{subfigure}
	\caption{Detecting two independent superfeatures in a combined image created from MNIST and FashionMNIST datasets.}
	\label{fig:sf}
\end{figure}

In addition to the above experiments to test the performance of the overall KED framework, here we present a special experiment to directly evaluate the superfeature construction method in Algorithm \ref{algo:sf}. We combine images and their labels from MNIST and FashionMNIST datasets to create a new dataset MNIST+FashionMNIST with 1568 features and 100 classes. For simplicity of illustration, let's say MNIST has 10 classes, namely `0',..., `9', and FashionMNIST has 10 classes, namely `A',..., `J'. Then the generated MNIST+FashionMNIST dataset has 100 classes, namely `0A',..., `9J'. In this combined dataset, we know the natural superfeature composition: there are two superfeatures, one containing MNIST features, and the other containing FashionMNIST features. In Figure \ref{fig:sf}, we show an example of the combined image and the superfeatures constructed by Algorithm \ref{algo:sf}. Note that this is different from image segmentation since our algorithm does not utilize pixel arrangement information and thus will discover the superfeatures even if the pixels of the combined image are randomly shuffled. Our experimental setup for this dataset is the same as that for MNIST in Section \ref{subsec:setup}. 

Table \ref{tab:comb} shows the distillation performance using the actual superfeatures, the superfeatures constructed by Algorithm \ref{algo:sf}, and two sets of randomly generated superfeatures. We observe that the superfeatures constructed by Algorithm \ref{algo:sf} achieve similar performance as the actual superfeatures. Furthermore, the superfeatures constructed by Algorithm \ref{algo:sf} provides significant performance gain over random groupings.

\begin{table*}[ht]
	\caption{Performance of Algorithm \ref{algo:sf} on MNIST+FashionMNIST combined dataset.}
	\label{tab:comb}
	\centering
	\begin{tabular}{cccccc}
		\toprule
		\multirow{2}{*}{Superfeature Design}	  & \multicolumn{2}{c}{Teacher} & \multicolumn{3}{c}{Student}\\ \cmidrule{2-6}
		& Black-box & Type-$M$ & No Distl. &  KD & KED \\	\midrule
		Actual  & \multirow{4}{*}{82.55\textpm0.05\%} & 85.83\textpm0.05\% & \multirow{4}{*}{75.09\textpm0.07\%} & \multirow{4}{*}{76.31\textpm0.06\%} & 82.82\textpm0.06\% \\
		Algorithm \ref{algo:sf}         &  & 85.31\textpm0.05\% &  &  & 81.89\textpm0.06\% \\
		Random 1                        &  & 82.14\textpm0.06\% &  &  & 77.28\textpm0.06\% \\
		Random 2                        &  & 82.06\textpm0.05\% &  &  & 76.57\textpm0.07\%  \\
		\bottomrule
	\end{tabular}
\end{table*}

\section{Conclusion}
\label{sec:conclusion}

In this work, we propose a new KED framework for training a low-complexity student model with knowledge transfer from a more powerful teacher. Unlike the conventional KD, under KED the teacher does not simply give predictions to the student but also explains those predictions. The proposed solution can be adapted to reduce the complexity of CNNs and to allow more effective distillation along with hidden-representation distillation methods as well as small training datasets. Our experimental results show that a KED teacher transfers more knowledge and can substantially improve student learning, leading to superior distillation performance in a wide variety of datasets and network architectures.





\ifCLASSOPTIONcaptionsoff
  \newpage
\fi



%
\bibliography{references}

\bibliographystyle{IEEEtran}

\end{document}